\documentclass[10pt,twocolumn,letterpaper]{article}

\usepackage[pagenumbers]{cvpr} 

\usepackage{graphicx}
\usepackage{amsmath}
\usepackage{amssymb}
\usepackage{booktabs}
\usepackage{bm}
\usepackage{standalone}

\usepackage{threeparttable}
\usepackage{multirow}

\usepackage{hyperref}
\hypersetup{pagebackref,breaklinks,colorlinks}

\usepackage[capitalize]{cleveref}
\crefname{section}{Sec.}{Secs.}
\Crefname{section}{Section}{Sections}
\Crefname{table}{Table}{Tables}
\crefname{table}{Tab.}{Tabs.}

\def\confName{CVPR}
\def\confYear{2022}

\def\ourmethod{NCINet}

\begin{document}

\title{Do learned representations respect causal relationships?}

\author{Lan Wang and Vishnu Naresh Boddeti\\
Michigan State University\\
{\tt\small {wanglan3,vishnu}@msu.edu}
}
\maketitle

\begin{abstract}
  Data often has many semantic attributes that are causally associated with each other. But do attribute-specific learned representations of data also respect the same causal relations? We answer this question in three steps. First, we introduce \ourmethod{}, an approach for observational causal discovery from high-dimensional data. It is trained purely on synthetically generated representations and can be applied to real representations, and is specifically designed to mitigate the domain gap between the two. Second, we apply \ourmethod{} to identify the causal relations between image representations of different pairs of attributes with known and unknown causal relations between the labels. For this purpose, we consider image representations learned for predicting attributes on the 3D Shapes, CelebA, and the CASIA-WebFace datasets, which we annotate with multiple multi-class attributes. Third, we analyze the effect on the underlying causal relation between learned representations induced by various design choices in representation learning. Our experiments indicate that (1) \ourmethod{} significantly outperforms existing observational causal discovery approaches for estimating the causal relation between pairs of random samples, both in the presence and absence of an unobserved confounder, (2) under controlled scenarios, learned representations can indeed satisfy the underlying causal relations between their respective labels, and (3) the causal relations are positively correlated with the predictive capability of the representations. Code and annotations are available at: \url{https://github.com/human-analysis/causal-relations-between-representations}.
\end{abstract}
\section{Introduction}
\label{sec:intro}

\begin{figure}[t]
    \centering
    \begin{subfigure}{0.45\textwidth}
        \centering
        \includegraphics[width=0.9\textwidth]{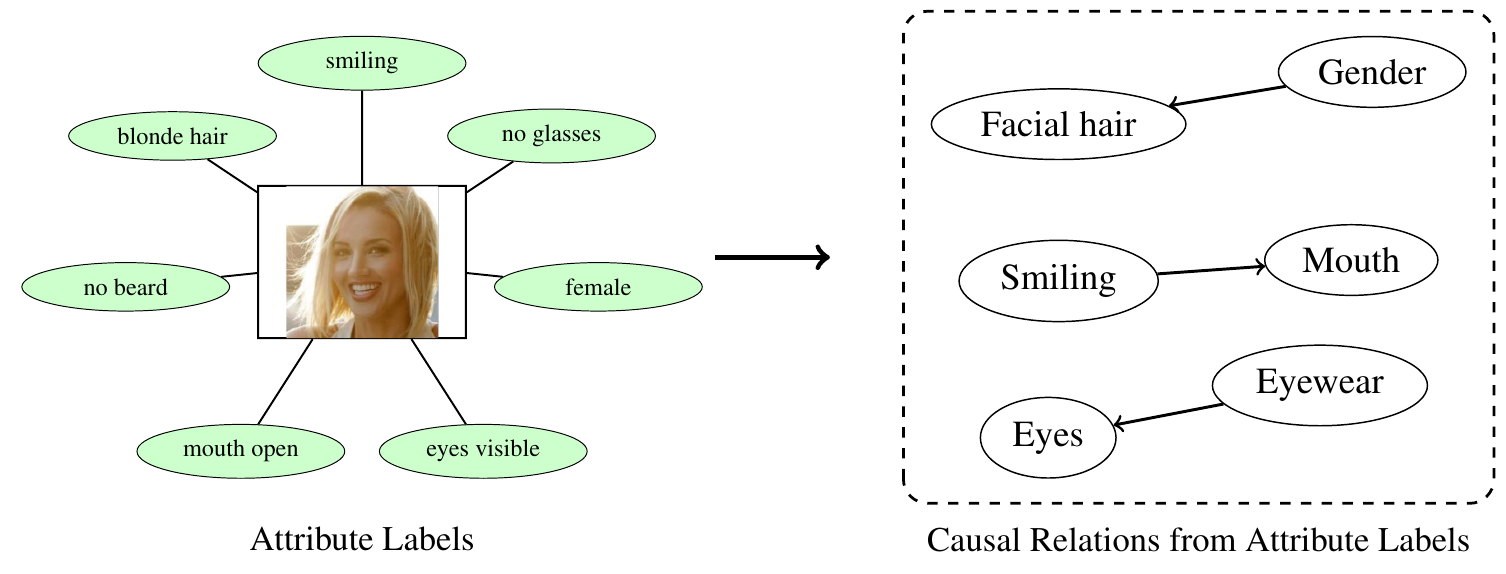}
        \caption{Illustration of causal relations between attribute labels\label{subfig:teaser1}}
    \end{subfigure}
    \begin{subfigure}{0.5\textwidth}
        \centering
        \includegraphics[width=0.9\textwidth]{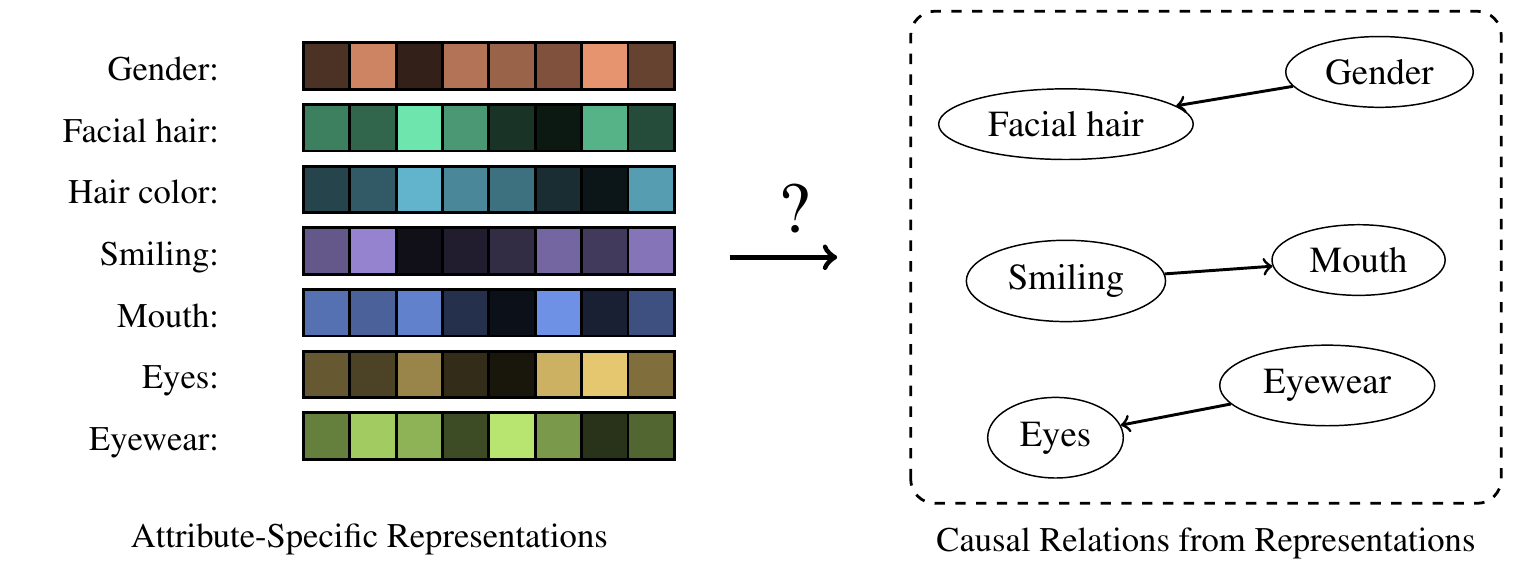}
        \caption{Causal relations between attribute-specific learned representations\label{subfig:teaser2}}
    \end{subfigure}
    \caption{Visual data may have multiple causally associated attributes. The goal of this paper is to determine whether attribute-specific learned representations respect the underlying causal relationships between the attributes? And if so, to what extent?\label{fig:teaser}\vspace{-0.5cm}}
\end{figure}

Consider the face image in Fig.~\ref{subfig:teaser1}. Automated face analysis systems typically involve extracting semantic attributes from the face. These attributes are often related through an underlying causal mechanism governing the relations between them. Modern computer vision systems excel at predicting such attributes by learning from large-scale annotated datasets. This is achieved by learning compact attribute-specific representations of the image from which the attribute prediction is made. This setting naturally raises the following questions (Fig. \ref{subfig:teaser2}): (1) \emph{Can we estimate the causal relations between high-dimensional representations purely from observational data with high accuracy?}, (2) \emph{Do the learned attribute-specific representations also satisfy the same underlying causal relations, and if so to what extent?}, and finally (3) \emph{How are the causal relations affected by factors such as the extent of training, overfitting, network architecture, etc.} Answering these questions is the primary goal of this paper.

Our work is motivated by the empirical observation that modern representation learning algorithms are inclined to uncontrollably absorb all correlations in the data~\cite{roy2019mitigating}. Consequently, while such systems have exhibited significantly improved empirical performance across many applications, it has also led to unintended consequences, ranging from bias against demographic groups \cite{blobaum2018cause} to loss of privacy by extracting and leaking sensitive information \cite{pittaluga2019revealing}. Identifying the causal relations between representations can help mitigate the deleterious effects of spurious correlations. With the proliferation of computer vision systems that employ such representations, it is imperative to devise tools to discover the causal relations given a set of representations.

Discovering causal relations from learned representations poses two main challenges. First, causal discovery typically involves interventions \cite{pearl2000causality} on the data which are either difficult or impossible on the observational representation space. For instance, in the image space, interventions may be possible during the image acquisition process for certain attributes such as hair color, eyeglasses, etc. Such interventions are, perhaps, not possible for attributes such as gender or ethnicity. On the other hand, it is not apparent how to intervene on any of these attributes directly in the representation space. Second, causal discovery methods, for pairs or whole graphs, are typically evaluated on small-scale low-dimensional datasets with multiple related attributes. However, there are no large-scale image datasets labeled with multiple causally associated attributes, nor are there any standardized protocols for evaluating the effectiveness of causal discovery methods on learned representations. While existing datasets such as MSCOCO~\cite{lin2014microsoft} and CelebA~\cite{liu2015deep} are labeled with multiple attributes they are either not causally related to each other (e.g., MSCOCO) or only have binary labels that suffer from severe class imbalance (e.g., CelebA).

To mitigate these challenges; (1) We propose Neural Causal Inference Net (\ourmethod{}) -- a learning-based approach for observational causal discovery from high-dimensional representations, both in the presence and absence of a confounder. \ourmethod{} is trained on a custom synthetic dataset of representations generated through a known causal mechanism. And, to ensure that it generalizes to real representations with complex causal relations we, (a) incorporate a diverse set of function classes with varying complexity into the data generating mechanism, and (b) introduce a learning objective that is explicitly designed to encourage domain generalization. (2) We develop an experimental protocol where, (a) existing datasets can be controllably resampled to induce a desired known causal relationship between the attribute labels, (b) learn attribute representations from the resampled data and infer the causal relations between them. We adopt three image datasets, namely, 3D Shapes dataset~\cite{3dshapes18}, CelebA~\cite{liu2015deep} and CASIA WebFace~\cite{yi2014learning}, where we annotate the latter with multiple multi-label attributes.

\noindent\textbf{Contributions:} First, we propose a learning-based tool, \ourmethod{}, for causal discovery from high-dimensional observational data, both in the presence and absence of a confounder. Numerical experiments on both synthetic and real-world data causal discovery problems indicate that \ourmethod{} exhibits significantly better causal inference generalization than existing approaches. Second, we employ \ourmethod{} for causal inference on attribute-specific learned representations and make the following observations; (1) Learned attribute-specific representations \emph{do} satisfy the same causal relations between the corresponding attribute labels under controlled scenarios with high causal strength. (2) The causal consistency is highly correlated with the predictive capability of the attribute classifiers (e.g., causal consistency degrades with overfitting).
\section{Related Work}

\begin{figure*}[ht]
\definecolor{darkgreen}{rgb}{0.4, 0.84, 0.2}
    \centering
    \begin{subfigure}{0.135\textwidth}
        \centering
        \includegraphics[scale=0.8]{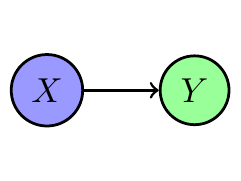}
         \caption{$\mathcal{G}_1$, Label: 1}
    \end{subfigure}
    \begin{subfigure}{0.135\textwidth}
        \centering
        \includegraphics[scale=0.8]{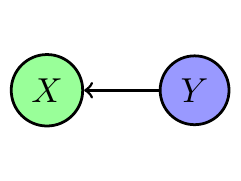}
        \caption{$\mathcal{G}_2$, Label: 2}
    \end{subfigure}
    \begin{subfigure}{0.135\textwidth}
        \centering
        \includegraphics[scale=0.8]{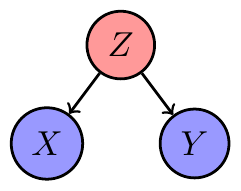}
        \caption{$\mathcal{G}_3$, Label: 0}
    \end{subfigure}
    \begin{subfigure}{0.135\textwidth}
        \centering
        \includegraphics[scale=0.8]{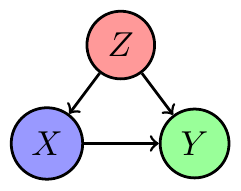}
        \caption{$\mathcal{G}_4$, Label: 1}
    \end{subfigure}
    \begin{subfigure}{0.135\textwidth}
        \centering
        \includegraphics[scale=0.8]{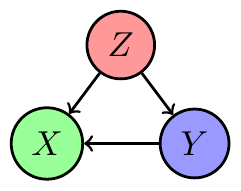}
        \caption{$\mathcal{G}_5$, Label: 2}
    \end{subfigure}
    \begin{subfigure}{0.135\textwidth}
        \centering
        \includegraphics[scale=0.8]{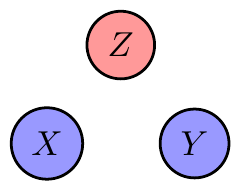}
        \caption{$\mathcal{G}_6$, Label: 0}
    \end{subfigure}
    \begin{subfigure}{0.135\textwidth}
        \centering
        \includegraphics[scale=0.8]{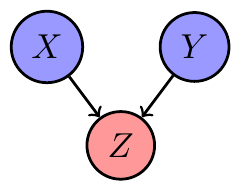}
        \caption{$\mathcal{G}_7$, Label: 0}
    \end{subfigure}
    \caption{All possible causal relations between pairs of random variables. A node in {\color{blue} blue} denotes cause, {\color{darkgreen} green} denotes effect, and {\color{red} red} denotes confounder or common effect ($Z$). We only consider scenarios where we observe $X$ and $Y$, but not $Z$. As such, the graphs represent three different causal relations, (i) Label 1: causal relation ($X \rightarrow Y$); (ii) Label 2: anti-causal relation ($X \leftarrow Y$); (iii) Label 0: $X$ and $Y$ are unassociated. Note that since $Z$ is not observed, $\mathcal{G}_7$ is equivalent to $\mathcal{G}_6$ and can thus be ignored. \label{fig:causalclass}\vspace{-0.5cm}}
\end{figure*}

\noindent\textbf{Representation Learning:} The quest to develop image representations that are simultaneously robust and discriminative has led to extensive research on this topic. Amongst the earliest learning-based approaches, Turk and Pentland proposed Eigenfaces \cite{turk1991face} that relied on principal component analysis (PCA) of data. Later on, integrated and high-dimensional spatially local features became prevalent for image recognition, notable examples include local binary patterns (LBP) \cite{ahonen2004face}, scale-invariant feature transform (SIFT) \cite{lowe1999object} and histogram of oriented gradients (HoG) \cite{dalal2005histograms}. In contrast to these hand-designed representations, the past decade has witnessed the development of end-to-end representation learning systems. Representations learned from supervised learning \cite{he2016identity,szegedy2017inception,liu2017sphereface}, disentangled learning \cite{higgins2016beta,chen2016infogan,tran2017disentangled,kim2018disentangling,chen2018isolating}, and most recently self-supervised learning \cite{doersch2015unsupervised,noroozi2016unsupervised,gidaris2018unsupervised,oord2018representation,chen2020simple} now typify modern image representations. The goal of these approaches is to learn universal representations that generalize well across arbitrary tasks. Hence, they are inclined to uncontrollably learn all contextual correlations in data. Our goal in this paper is to verify whether learned representations retain the underlying causal relations of the data generating process.

\vspace{3pt}
\noindent\textbf{Causal Inference:} Randomized controlled experiments are the gold standard of causal inference. However, in many computer vision applications, we cannot control the image formation process rendering such experiments infeasible. Concurrently, a plethora of approaches have been proposed for causal discovery purely from observational data under two main settings, full graph or pairwise. Estimating the full causal graph has been thoroughly studied, both using learning-based \cite{buhlmann2014cam,aragam2015concave, kalainathan2018structural,bengio2019meta,ke2019learning} and non-learning-based \cite{pearl2009causality,chickering2002optimal,spirtes2000causation, kalisch2007estimating} approaches.

In this paper, we restrict our focus to causal discovery for the particular case where we have access to only two random variables at a time, both in the presence or absence of an unobserved confounder. Significant efforts have also been devoted to this problem under different scenarios. These include, comparison of information entropy for discrete variables \cite{kocaoglu2016entropic,kocaoglu2018entropic}, neural causal methods \cite{lopez2017discovering, louizos2017causal, goudet2018learning}, comparison of noise statistics in causal and anti-causal directions \cite{hoyer2009nonlinear,mooij2016distinguishing}, comparison of regression errors between the causal and anti-causal directions \cite{blobaum2018cause}, comparison of Kolmogorov complexity \cite{vreeken2015causal,budhathoki2018origo}, building classification and regression trees \cite{marx2018causal}, analyzing conditional distributions \cite{kalainathan2019discriminant,fonollosa2019conditional} and many more \cite{janzing2009telling,peters2015causal,lopez2015towards}. A majority of the aforementioned methods have been designed and applied to low-dimensional variables, except for \cite{vreeken2015causal, budhathoki2018origo,marx2018causal,janzing2009telling}.

Within the broader context of computer vision, there is growing interest in causal discovery \cite{lopez2017discovering}, causal data generation \cite{kocaoglu2017causalgan}, incorporating causal concepts within scene understanding systems \cite{zhang2020causal,tang2020unbiased,wang2020visual,yang2021causal}, domain adaptation\cite{yue2021transporting}, and debiasing \cite{wang2021causal}.

In this paper, the proposed  \ourmethod{} is a neural causal inference method that is tailored for high-dimensional variables. It incorporates, (1) direct supervision through causal labels, indirect supervision by comparing regression errors in the causal and anti-causal direction and an adversarial loss to encourage domain generalization, and (2) in contrast to all existing approaches, our model is trained to infer causal relations from all possible (see Fig. \ref{fig:causalclass}) pairwise cases, including in the presence and absence of an unobserved confounder.
\section{Causal Relations Between Representations}
First, we define the primary causal inference query that this paper seeks to answer i.e., \emph{``Do learned representations respect causal relationships?"}. Consider the graph $\mathcal{G}_1$ in Fig.~\ref{fig:causalclass}, which has two attributes $X$ and $Y$, where the causal relation between them is $X \rightarrow Y$. An image $\bm{I}$ is generated by an unknown stochastic function of these two attributes. Let $\bm{x}$ and $\bm{y}$ be high-dimensional attribute-specific representations learned for predicting labels $X$ and $Y$, respectively, from the corresponding images. The structural causal equations (SCEs)\footnote{The SCEs and the corresponding causal inference queries for the other pairwise causal relations in Fig.~\ref{fig:causalclass} can be defined similarly.} that characterize this process are:
\begin{equation}
    \begin{aligned}
        a_x \sim P_c(X) \quad  a_y \sim P_e(Y|X=a_x) \\
        \bm{I} = g(a_x,a_y,\epsilon) \\
        \bm{x} = h_X(\bm{I};\bm{\theta}_X) \quad \bm{y} = h_Y(\bm{I};\bm{\theta}_Y) \\
    \end{aligned}
\end{equation}
\noindent where $a_x$ and $a_y$ are sampled attribute instances, $\epsilon$ is a noise variable which is independent of both $X$ and $Y$ and $h_X(\cdot;\bm{\theta}_X)$ and $h_Y(\cdot;\bm{\theta}_Y)$ are the encoders that extract the attribute-specific representations for $X$ and $Y$, respectively. Under this model, given the distribution of features $\bm{x} \sim P(\bm{z}_x)$ and $\bm{y} \sim P(\bm{z}_y)$ for the two attributes, we seek to determine whether the attribute-specific representations also follow the same underlying causal relations, i.e., \emph{is $\bm{z}_x \rightarrow \bm{z}_y$?}. The association between these learned attribute features can be well approximated as a post nonlinear causal model (PNL)~\cite{zhang2010distinguishing},
\begin{equation}
    \bm{z}_y = f_2(f_1(\bm{z}_x)+\bm{\epsilon}) 
\end{equation}
\noindent where $f_2$ and $f_1$ are non-linear functions with $f_2$ being continuous and invertible, and $\bm{\epsilon}$ is a noise variable such that $\bm{e} \perp\!\!\!\!\perp \bm{z}_x$. The identifiability of the PNL model from observational data was established by Zhang and Hyv{\"a}rinen \cite{zhang2012identifiability}. Conceptually, the key idea is that the distribution $P(\bm{z}_y|\bm{z}_x)$ in the causal direction is \emph{``less complex"} than that in the anti-causal direction. \ourmethod{}, the proposed causal inference approach, is designed to exploit such disparity. 

We note that, in the absence of strong assumptions, direct causal relation is indistinguishable from that induced by latent confounding. However, we take inspiration from the ability of humans to accurately infer causal relations only from observations in many cases, and seek to unveil causal motifs between the representations directly from samples.

\section{Observational Causal Discovery Problem}
Learning-based observational causal discovery considers a dataset $\mathcal{S}$ of $n$ \emph{observational samples},
\begin{equation}
    \mathcal{S} = \{S_i\}_{i=1}^n = \{(\bm{x}_j,\bm{y}_j)_{j=1}^{m_i}\}_{i=1}^n \sim P(\bm{x},\bm{y})
\end{equation}
\noindent where each sample $S_i$ is itself a dataset of $m_i$ representation pairs $\{(\bm{x}_1,\bm{y}_1),\dots,(\bm{x}_{m_i},\bm{y}_{m_i})\}$, $\bm{x}\in\mathbb{R}^{d_x}$ and $\bm{y}\in\mathbb{R}^{d_y}$ are the learned representations corresponding to predicting $X$ and $Y$ respectively, and $P(\bm{x},\bm{y})$ is the joint distribution of the two representations. The joint distribution $P(X,Y)$ can represent different causal relations as shown in Fig.~\ref{fig:causalclass}, namely, (i) causal class ($X \rightarrow Y$); (ii) anti-causal class ($X \leftarrow Y$); (iii) X and Y are unassociated, both in the absence and presence of an unobserved confounder $Z$.

\begin{figure*}[!ht]
    \centering
    \includegraphics[width=0.9\textwidth]{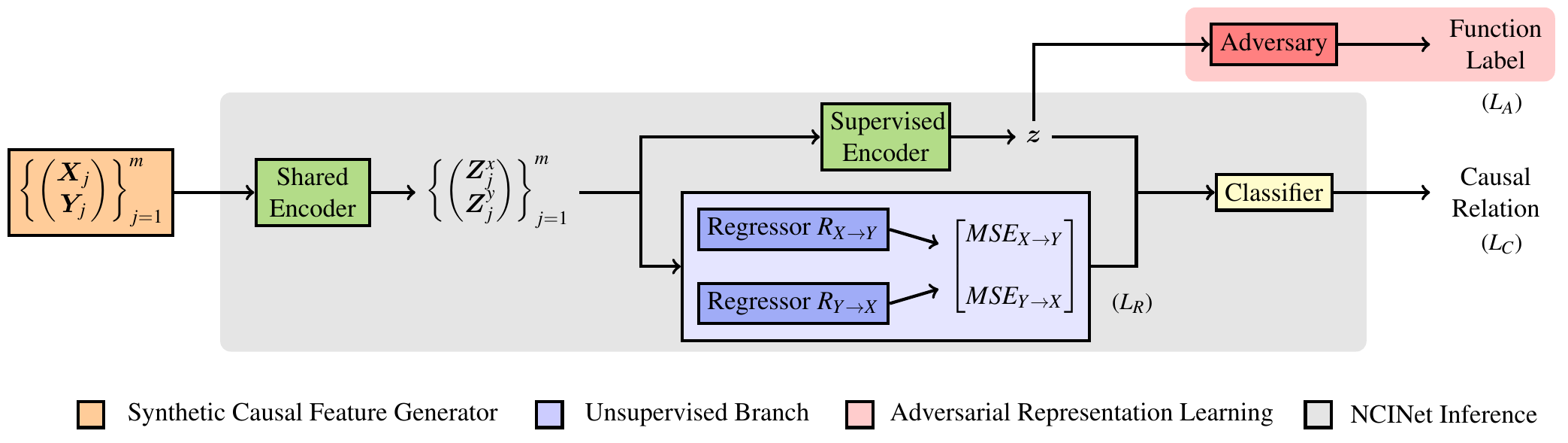}
    \caption{\textbf{Overview:} Schematic illustration of Neural Causal Inference Net (NCINet). It comprises of, (1) a shared encoder that maps representations to a common space, (2) a supervised encoder that extracts a representation $\bm{z}$ from the common space, (3) a causal regression branch that compares the regression errors in the causal and anti-causal direction, (4) an adversary that seeks to extract the function label, and (5) a fusion module that combines information from the two branches and predicts the causal relationship. See text for more details.\label{fig:framework}\vspace{-0.5cm}}
\end{figure*}

The key idea of learning-based causal discovery is to exploit the many manifestations of \emph{causal footprints} often present in real-world observational data \cite{peters2014causal}. For example, oftentimes the functional relationships in the causal direction are ``simpler" than those in the anti-causal direction. Unsupervised methods exploit such \emph{causal signals} either by measuring the complexity of the causal and anti-causal functions \cite{budhathoki2018origo}, the entropy of causal and anti-causal factorizations of the joint distribution \cite{kocaoglu2016entropic} or comparing regression errors in the causal and anti-causal direction \cite{blobaum2018cause}.

Going beyond a specific type of \emph{causal footprint}, supervised methods seek to exploit any and all possible \emph{causal signals} in the observational data by learning to directly predict causal labels from the observation dataset $\mathcal{S}$. \emph{Neural causal models} such as NCC \cite{lopez2017discovering}, GNN \cite{goudet2018learning} and CEVAE \cite{louizos2017causal} are a special class of supervised approaches that leverage neural networks based classifiers.

Although both supervised and unsupervised approaches are based on the same principle -- namely, \emph{exploiting causal footprints} -- they differ in one key aspect. Unlike the unsupervised methods, the supervised approaches need ground truth causal labels to train the causal classifier. However, in most real-world scenarios the ground truth causal graph is unknown. Therefore, the supervised methods are typically trained purely on synthetically generated data and hence suffer from a synthetic-to-real domain generalization gap. Unsupervised methods on the other hand can be applied directly to the observational data of interest and hence are agnostic to the data domain. However, unlike the supervised approaches, the unsupervised methods like RECI~\cite{blobaum2018cause} exploit only one type of \emph{causal footprint} at a time, e.g., regression errors between the causal and anti-causal directions.

\section{Neural Causal Inference Network}

Neural Causal Inference Network (\ourmethod{}) is a neural causal model for observational causal discovery. Given a pair of high-dimensional attribute-specific representations $S = \{\bm{x}_j,\bm{y}_j\}_{j=1}^m$, we seek to determine one of three causal relations, $X\rightarrow Y$, $X\leftarrow Y$ or $X$ is unassociated with $Y$. Fig.~\ref{fig:framework} shows a pictorial overview of \ourmethod{} along with a causal data generation process that is customized for high-dimensional signals.

Our entire solution is motivated from three perspectives: (1) \textbf{Modeling:} As described in the previous section, the supervised and unsupervised models have complementary advantages and limitations. Therefore we incorporate both of them into \ourmethod{} to make a final prediction. (2) \textbf{Data:} Semantic image attributes (e.g., facial features) span the whole spectrum of pairwise causal relations illustrated in Fig.~\ref{fig:causalclass}. However, existing supervised and unsupervised learning-based causal discovery methods only consider a subset of these relations (ignoring either the independence class or the unobserved confounder) and are designed for low-dimensional signals, and therefore sub-optimal or insufficient for our purpose. Therefore, we design a synthetic data generation process for obtaining high-dimensional features spanning all possible pairwise causal relations. (3) \textbf{Generalization:} For \ourmethod{} to generalize from the synthetic training data to real representations we adopt two strategies. First, the synthetic feature generation process includes an ensemble of linear and non-linear causal functions. Second, we employ an adversarial loss to debias the prediction w.r.t the choice of functional classes in the synthetic training data.

\ourmethod{} comprises of five components: shared encoder, supervised encoder, causal regression branch, adversary, and classifier. These components are described below.

\noindent\textbf{Encoders:} There are two encoders, a shared encoder that maps the pair of representations into an intermediate representation $(\bm{z}_x, \bm{z}_y) =\left(E_{SE}(\bm{x}),E_{SE}(\bm{y})\right)$, and a supervised encoder that extracts features for the final classifier. The latter encoder acts on the concatenated features $\begin{bmatrix}\bm{z}_x & \bm{z}_y\end{bmatrix}^T$ and extracts a representation that is average pooled over the $m$ samples in the representation. The resulting feature is denoted as $\bm{z}$ in Fig.~\ref{fig:framework}.

\vspace{2pt}
\noindent\textbf{Causal Regression:\label{sec:Regression}} The Causal Regression branch of \ourmethod{} is inspired by the asymmetry idea proposed by \cite{blobaum2018cause}, wherein the mean squared error (MSE) of prediction is smaller in the causal direction in comparison to the anti-causal direction, i.e., 
\begin{equation}
    \mathbb{E}[(E-\phi(C))^2] \leq \mathbb{E}[(C-\psi(E))^2],
\end{equation}
where $C$ is the cause and $E$ is the effect, $\phi$ is the regressor that minimizes the MSE when predicting $E$ from $C$, and $\psi$ is the regressor that minimizes the MSE when predicting $C$ from $E$. Therefore, the causal relation can be estimated by comparing the two regression errors. An attractive property of this idea is its inherent ability to generalize to unseen causal data generating functions classes by virtue of being unsupervised and not requiring any learning.

The causal regression branch of \ourmethod{} adopt ridge regressors that operates on the intermediate embeddings ($\bm{z}_x,\bm{z}_y$). The causal regressor $R_{X \rightarrow Y}: \bm{z}_x \mapsto \bm{y}$ minimizes the MSE, $\frac{1}{m}\sum_{j=1}^m\|\hat{\bm{y}}_j - \bm{y}_j\|_2^2$, between the prediction $\hat{\bm{y}}=\left(\bm{z}_x^T\bm{z}_x+\lambda\bm{I}\right)^{-1}\bm{z}_x\bm{y}$ and ground truth input $\bm{y}$. Similarly, the anti-causal regressor $R_{X \leftarrow Y}: \bm{z}_y \mapsto \bm{x}$ minimizes the MSE, $\frac{1}{m}\sum_{j=1}^m\|\hat{\bm{x}}_j - \bm{x}_j\|_2^2$, between the prediction $\hat{\bm{x}}=\left(\bm{z}_y^T\bm{z}_y+\lambda\bm{I}\right)^{-1}\bm{z}_y\bm{x}$ and ground truth input $\bm{x}$. The two regressors $R_{X\rightarrow Y}$ and $R_{X \leftarrow Y}$ are trained end-to-end (i.e., we backpropagate through the closed-form ridge regressor solution) along with the rest of the components in \ourmethod{}. Therefore, the loss from the regression branch is,
\begin{equation}
L_R = \frac{1}{m}\sum_{i=1}^m\|\hat{\bm{y}}_i - \bm{y}_i\|_2^2 + \frac{1}{m}\sum_{i=1}^m\|\hat{\bm{x}}_i - \bm{x}_i\|_2^2
\end{equation}

\vspace{2pt}
\noindent\textbf{Adversarial Loss:} The features $\bm{z}$ extracted from the supervised encoder potentially still contain information specific to the function class that generated the synthetic features. However, the generalization performance of \ourmethod{} may be hampered if the downstream classifier exploits any spurious correlation between the function class-specific information and the ground truth causal relations. Therefore, we measure the amount of information in $\bm{z}$ about the function class through an adversary and minimize it. This type of adversary is typically modeled as a neural network and optimized via min-max optimization, which can be unstable in practice~\cite{jin2019local,daskalakis2018limit}. For ease of optimization we instead model the adversary by a kernel ridge-regressor which admits a closed-form solution $\hat{\bm{y}}_f = \bm{K}\left(\bm{K} + \beta \bm{I}\right)^{-1}\bm{y}_f$, where $\bm{y}_f$ is the one-hot vector representing the function class of the synthetic data, $\beta$ is a regularization, and $\bm{K}$ is a kernel matrix computed from the features $\bm{z}$. The loss from the adversary which we backpropagate through is, 
\begin{equation}
    L_A = -\|\bm{y}_f - \hat{\bm{y}}_f\|^2_2 = -\|\bm{y}_f - \bm{K}\left(\bm{K} + \beta \bm{I}\right)^{-1}\bm{y}_f\|_2^2
\end{equation}

\vspace{2pt}
\noindent\textbf{Classifier:} Finally, the supervised classifier concatenates the features $\bm{z}$ from the supervised encoder with the output $\begin{bmatrix}MSE_{X\rightarrow Y} & MSE_{Y\rightarrow X} & \frac{min(MSE_{X\rightarrow Y}, MSE_{Y\rightarrow X})}{max(MSE_{X\rightarrow Y}, MSE_{Y\rightarrow X}}) \end{bmatrix}^T$ of the causal regressors, and categorizes the causal relations into three classes as follows.
\begin{equation}
l = \begin{cases} 0 &\mbox{if } X \mbox{ unassociated with } Y \mbox{ i.e., no causal relation }\\
1 & \mbox{if } X \rightarrow Y \\
2 & \mbox{if } X \leftarrow Y \end{cases} \nonumber
\end{equation}
We note that this is unlike existing methods for causal inference such as NCC~\cite{lopez2017discovering}, RECI~\cite{blobaum2018cause}, etc., which only categorize the causal relations into two categories, namely causal and anti-causal. However, in many practical scenarios, the image attributes and their corresponding representations could be very weak as we discuss in Section~\ref{sec:unknown}. The classifier minimizes the cross-entropy loss $L_C$ between its prediction and the ground truth causal relations $l$.

All the components of \ourmethod{} are trained end-to-end by simultaneously optimizing all the intermediate losses i.e., $Loss = L_C + L_R + \lambda L_A$,
where $\lambda$ is the weight associated with the adversarial loss. The classifier will learn to exploit all the \emph{causal footprints} in the data aided by the causal regressors and the adversary. The features from the causal regressors help exploit the \emph{causal footprint} corresponding to the difference in regression errors, while the adversary helps the classifier to reduce the synthetic-to-real domain generalization gap. Interaction between the regressors, adversary, and the final classifier is induced by the common intermediate representation space $(\bm{z}_x,\bm{z}_y)$, on which all of them operate.
\section{Experiments: Neural Causal Inference\label{sec:experiments}}

In this section, we evaluate the performance and generalization ability of \ourmethod{} in comparison to existing baseline methods on synthetically generated high-dimensional representations with known causal relations.

\vspace{2pt}
\noindent\textbf{Data and Training:}
The lack of large-scale datasets with ground truth causal labels precludes causal discovery models from being trained on real-world observational data. Therefore, it is standard practice to train and evaluate causal discovery models on synthetic observational data. Models trained in this manner can now be applied directly to real-world observational data. Synthetic data generation typically follows the additive noise model \cite{pearl2000models}, where an effect variable is obtained as a function of causal variable and perturbed with independent additive noise. We adopt the same additive noise model as our causal mechanism. 

To improve generalization capability, we diversify the synthetic training data. Specifically, we adopt an ensemble of different high-dimensional causal functions including, Linear, Hadamard, Bilinear, Cubic Spline, and Neural Networks. See the supplementary material for more details. In each training epoch, we generate 1000 samples, where each data sample consists of 100 feature pairs (i.e., $m=100$) by randomly sampling one of the causal functions and their respective parameters. Integrating data generation into the training process ensures that the models learn from an infinite stream of non-repeating data.

We generate the pairs of representations $(\bm{x}, \bm{y})$ via ancestral sampling. For example, in the case of $\mathcal{G}_1$ where $\bm{x}\rightarrow \bm{y}$, the synthetic representations are generated as follows, $P(\bm{w})\rightarrow P(\bm{x}|\bm{w}) \rightarrow P(\bm{y}|\bm{x},\bm{w})$, where $\bm{w}$ accounts for all the unobserved confounders. More details can be found in the supplementary material.

\vspace{2pt}
\noindent\textbf{Baselines:}
We consider four baseline methods, ANM \cite{hoyer2009nonlinear}, Bivariate Fit (BFit) \cite{kalainathan2019causal}, NCC \cite{lopez2017discovering} and RECI \cite{blobaum2018cause}.  These methods, however, were originally designed for causal inference on one-dimensional variables and to distinguish between causal and anti-causal directions. Therefore, we extend them to high-dimensional data, as well as to distinguish between causal direction, anti-causal direction, and no causal relation. Specifically, for NCC, we concatenate high-dimensional features $\bm{x}$ and $\bm{y}$ as the input to the network and change the output layer to three classes. For the unsupervised methods, ANM, BFit, and RECI, we regress directly on high-dimensional features $\bm{x}$ and $\bm{y}$ as required. Since these methods are score based i.e. $score > 0$ represents the causal direction and $score < 0$ represents the anti-causal direction we introduce an additional threshold to identify the no causal relation case i.e. if $|score| < threshold$. We use a separate validation set to determine the optimal threshold for each unsupervised method.

\vspace{2pt}
\noindent\textbf{Generalization Results:} To evaluate the performance of the models and their generalization ability, we adopt a leave-one-function out evaluation protocol. For evaluating each causal function, we train the models using data generated by all the other causal functions across all the causal graphs in Fig. \ref{fig:causalclass}. The results are shown in Table~\ref{tab:generalization} for 8 dimensional representations. We observe that overall \ourmethod{} outperforms all the baselines.

\begin{table}[htbp]
  \centering
  \caption{Leave-one-function out mean accuracy (\%) of five runs on different causal functions with 8 dimensional features (see supplementary for more details). Best results are in bold.\label{tab:generalization}}
    \scalebox{0.65}{
    \begin{tabular}{lccccccc}
    \toprule
    Methods && \multicolumn{1}{l}{Linear} & \multicolumn{1}{l}{Hadamard} &  \multicolumn{1}{l}{Bilinear} & \multicolumn{1}{l}{Cubic Spline} & \multicolumn{1}{l}{NN} & \multicolumn{1}{l}{Average} \\
    \cline{1-1} \cline{3-8} \\
    ANM~\cite{hoyer2009nonlinear}   && 31.87  & 32.49 & 32.94 & 33.66 & 33.08 & 32.81  \\
    Bfit~\cite{kalainathan2019causal}  && 34.89 & 54.76  & 53.69  &\textbf{77.79} & 38.26  &51.88 \\
    NCC~\cite{lopez2017discovering}   && 52.64 & 83.93 & 85.66  & 77.03 & 56.56  & 71.16 \\
    RECI~\cite{blobaum2018cause}  && 42.73   & \textbf{89.66 }& \textbf{92.02}& 71.49 & 60.23 & 71.43 \\
    \midrule
   \ourmethod{} && \textbf{64.16 } & 81.13  & 89.73  & 71.33  &  \textbf{69.53  } & \textbf{75.17}\\
    \bottomrule
    \end{tabular}}%
\end{table}

\section{Causal Inference on Learned Features\label{sec:learned-features}}
Our goal in this section is to estimate the causal relation between attribute-specific learned representations and verify if it is consistent with the causal relations between the corresponding labels. Furthermore, we would like to perform this analysis for all the different types of causal relations shown in Fig.~\ref{fig:causalclass}. However, real-world datasets only provide a fixed collection of images and their corresponding labels, and as such do not afford any explicit control over the type and strength of causal relations between the attributes. To overcome the aforementioned limitations we consider causal inference on learned representations under two scenarios where the causal relations between the attributes are known and unknown.

\noindent\textbf{Datasets:} We consider three image datasets with multiple multi-label attributes; (1) 3D Shapes~\cite{3dshapes18} which contains 480,000 images. Each image is generated from six latent factors (floor hue, wall hue, object hue, scale, shape, and orientation) which serve as our image attributes. (2) CASIA WebFace~\cite{yi2014learning} which contains 494,414 images of 10,575 classes. While the dataset does not come with attribute annotations, we annotate each image with eight multi-label attributes\footnote{The choice of attributes and labels for each may arguably still not fully reflect the real-world. Nonetheless, we believe this dataset could be a valuable resource for causal analysis tasks.} (color of hair, visibility of eyes, type of eyewear, facial hair, whether the mouth is open, smiling or not, wearing a hat, visibility of forehead, and gender). The annotations for this dataset have been made \href{https://github.com/human-analysis/causal-relations-between-representations}{publicly available} to the research community. (3) CelebA~\cite{liu2015deep} which contains 202,599 images, each of which is annotated with 40 binary attributes. However, this dataset suffers from severe class imbalance across most attributes. Experimental results on this dataset can be found in the supplementary material.

\vspace{2pt}
\noindent\textbf{Learning Attribute-Specific Representations:} We learn attribute-specific representations by learning attribute predictors for each attribute. For predicting the attributes, we use a 5-layer CNN and ResNet-18~\cite{he2016identity} for the 3D Shapes and CASIA-WebFace datasets, respectively. The attribute predictors are optimized using AdamW~\cite{loshchilov2018fixing} with a learning rate of $5\times 10^{-4}$ and a weight decay of $5\times 10^{-4}$. Upon convergence of the attribute predictor training, we use the trained model to extract representations from the layer before the linear classifier at the end.

\vspace{2pt}
\noindent\textbf{Causal Inference Baselines:} To infer the causal relations between the learned attribute representations we apply \ourmethod{} and two other baselines NCC and RECI. Both \ourmethod{} and NCC are trained on the synthetic dataset as described in Section \ref{sec:experiments}.

\begin{figure*}[t]
    \centering
    \begin{subfigure}{\textwidth}
        \centering
        \vspace{- 0.5cm}
        \includegraphics[width=\textwidth]{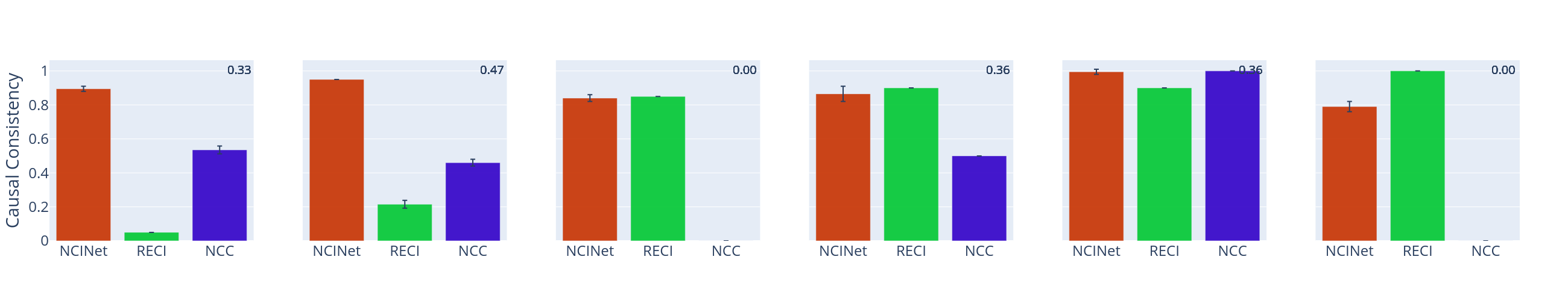}
    \end{subfigure}
    \vspace{- 0.5cm}
    \begin{subfigure}{\textwidth}
        \centering
        \vspace{- 0.2cm}
        \includegraphics[trim={0 0 0 3cm},clip,width=\textwidth]{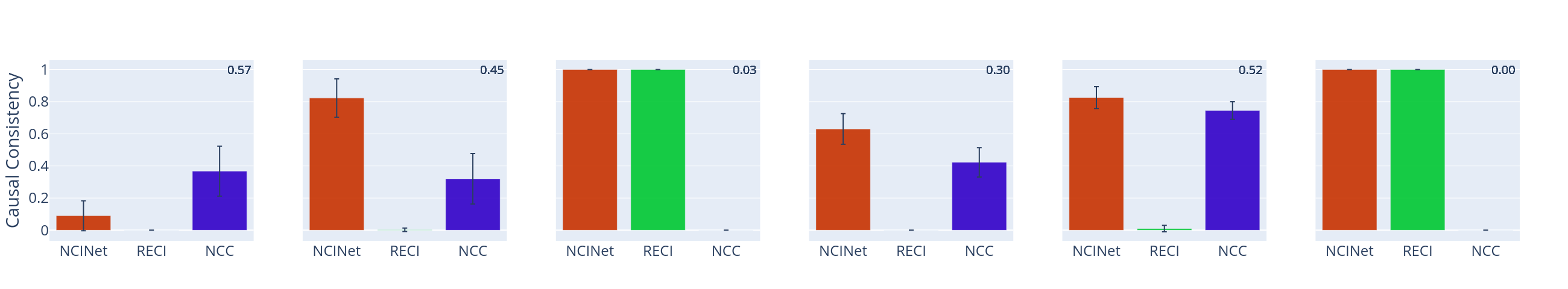}
    \end{subfigure}
    \caption{\emph{Causal consistency} between the labels and representations of attributes. The columns represent the different types of causal relations starting from $\mathcal{G}_1$ on the left and $\mathcal{G}_6$ on the right. The strength of the causal relation between the labels, estimated through \cite{kocaoglu2016entropic}, is shown in top right-hand corner of each subplot. (Top) 3D Shape and (Bottom) CASIA WebFace dataset.\label{fig:results-known}\vspace{-0.2cm}}
\end{figure*}

\subsection{Known Causal Relations Between Labels\label{sec:known}}
In this experiment, we resample the datasets to obtain samples with the desired type of causal relations. Consequently, in this scenario, the causal relationship between the attribute labels is known, and we seek to verify if the corresponding attribute-specific learned representations also satisfy the same causal relations. We note that, although the data generated in this way will not reflect the true underlying causal relations between the attributes, it nonetheless allows us to perform controlled experiments. We chose floor hue and wall hue as the attribute for 3D Shapes, and visibility of the forehead, and whether the mouth is open as the attributes for CASIA-WebFace. The choice of the attributes of CASIA-WebFace was motivated by the fact that these two attributes were the most sample balanced pair of attributes. For each type of causal relation in Fig.\ref{fig:causalclass}, we sample 2000/2000 and 8000/2000 images for training/testing on 3D Shapes and CASIA-WebFace, respectively.

\vspace{2pt}
\noindent\textbf{Generating Images with Causally Associated Attributes:} The data sampling process proceeds in two phases. In the first phase, we generate the attribute labels with known causal relations. We represent each type of causal graph via its corresponding Bayesian Network with hand-designed conditional probability tables. Then we sample batches of labels with known causal relations through Gibbs Sampling. To further ensure that the sampled labels correspond to the desired causal relation, we measure the strength of their causal relation through an entropic causal inference method \cite{kocaoglu2016entropic}. In the second phase, we sample images that conform to the sampled attribute labels. On the 3D Shapes dataset, to increase the diversity of the images and ensure that the representation learning task is sufficiently challenging, the images are corrupted with one of three types of noise, Gaussian, Shot, or Impulse.

\vspace{2pt}
\noindent\textbf{Results:} To measure the consistency between the causal relations of the labels and the causal relations of the respective representations, we introduce a new metric dubbed \emph{causal consistency}. For a given set of learned representations $\left(\bm{x}_j,\bm{y}_j\right)_{j=1}^m$, we split it into multiple non-overlapping subsets. The causal relation is estimated for each subset and we measure how many of them are consistent with the causal relation $l$ between the labels. Evaluating over multiple subsets serves to prevent outliers from severely affecting the causal inference estimates (see supplementary material for more details). Fig.~\ref{fig:results-known} shows the \emph{causal consistency} results across the different types of causal relations. We make the following observations: (1) In most cases, across both 3D Shapes and CASIA-WebFace, the causal relationship between the learned representations is highly consistent with that of the labels. This empirical evidence is encouraging since it suggests that representation learning algorithms are capable of mimicking the causal relations inherent to the training data. (2) In the controlled setting of this experiment, among the three causal inference methods, \ourmethod{} appears to provide more stable and consistent estimates of causal relations across the different causal graphs and datasets, followed by RECI and NCC. %

\subsection{Unknown Causal Relations Between Labels\label{sec:unknown}}
In this experiment, we consider the original CASIA-WebFace dataset as is, without any controlled sampling. We choose smiling or not and visibility of eyes as the two attributes to investigate. The attribute predictors are trained/validated on 10,000/10,000  randomly sampled images using a ResNet-18 architecture. Other training details are similar to the experiment in Section~\ref{sec:known}. Since the true causal relation between the labels is unknown, we use an entropic causal inference method \cite{kocaoglu2016entropic} to estimate it. While the causal relation between the labels suggests that smiling has an effect on the visibility of eyes its causal strength is very weak (0.23/0.20 for training/validation). The causal relation between the learned representations follow the same trend, with around 20\% of the representation subsets agreeing with smiling having an effect on the visibility of eyes, while 80\% of them suggest that there is no causal relation between the two attributes.

\section{Discussion\label{sec:discussion}}
This section analyzes the effect of various aspects of representation learning on the \emph{causal consistency} between the learned representations and the labels.
\begin{table}[htbp]
    \centering
    \caption{ Effect of Adversarial Debiasing on \ourmethod{} (one run)} \label{tab:ablation} \vspace{-0.25cm}
    \scalebox{0.7}{
    \begin{tabular}{lcccccc}
    \toprule
    \ourmethod{} & \multicolumn{1}{l}{Linear} & \multicolumn{1}{l}{Hadamard} & \multicolumn{1}{l}{Bilinear} & \multicolumn{1}{l}{Cubic spline} &\multicolumn{1}{l}{NN} & \multicolumn{1}{l}{Average} \\
    \midrule
        w/o Adv & 66.50 & 80.33 & 89.67 & 70.5 & 67.17 &  74.83\\
        w/ Adv & 66.67  & 80.50 & 90.17 & 71.00  & 68.33 &  75.33\\
    \bottomrule
    \end{tabular}}
\end{table}

\vspace{3pt}
\noindent\textbf{Effect of Adversarial Debiasing on \ourmethod{}:} We study the contribution of adversarial loss $L_A$ on the generalization performance of \ourmethod{}. Table~\ref{tab:ablation} reports the causal inference results on the synthetically generated representations in Section~\ref{sec:experiments}. Overall the adversarial loss aids in improving the generalization ability of \ourmethod{}, thereby validating our hypothesis that the features $\bm{z}$ from the supervised encoder still contains information specific to the function class.

\begin{figure}[!ht]
    \centering
    \includegraphics[scale=0.16]{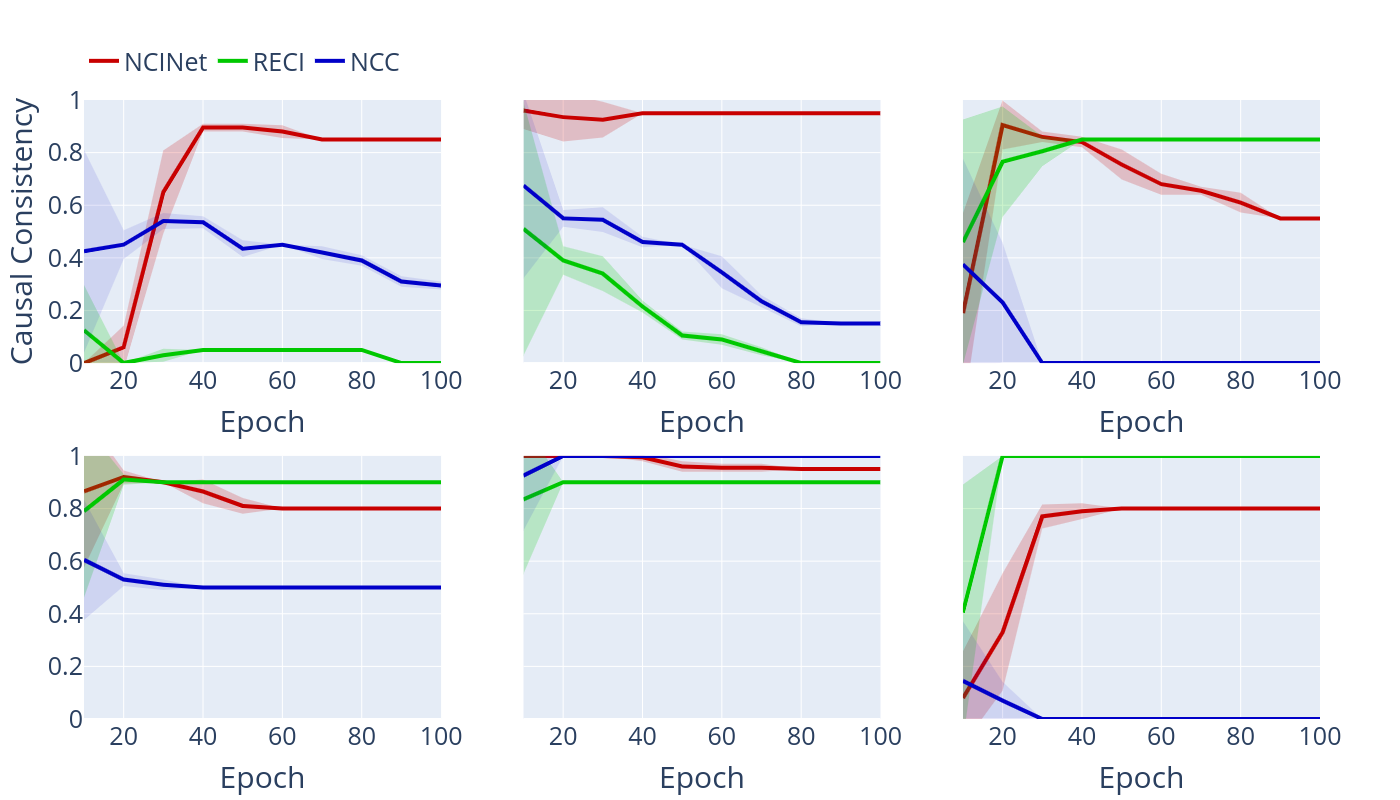}
    \caption{\emph{Causal consistency} as a function of training epochs for the six different types of causal relations. (Top) $\mathcal{G}_1$-$\mathcal{G}_3$ and (Bottom) $\mathcal{G}_4$-$\mathcal{G}_6$. Bands show standard deviation.\label{fig:epochs}}
\end{figure}

\vspace{3pt}
\noindent\textbf{Effect of Training Epochs:} Here we study the causal relations between the representations as a function of the training epochs. In this experiment, we test the causal consistencies of features extracted by models from each training epoch, and show the average value of every 10 epochs.
We hypothesize that as the attribute prediction performance of the representations improves, the causal relation between the representations will also become more consistent with the causal relation between the labels. Fig.~\ref{fig:epochs} shows the \emph{causal consistency} as a function of the training epochs. We make three observations: (1) as training progresses, the \emph{causal consistency} of \ourmethod{} improves and remains stable which is consistent with our hypothesis. (2) Although RECI fails on $\mathcal{G}_1$ and $\mathcal{G}_2$, it follows the same trend in the other cases. (3) NCC is prone to failure in no causal relation data.

\begin{figure}[!ht]
    \centering
    \includegraphics[width=0.5\textwidth]{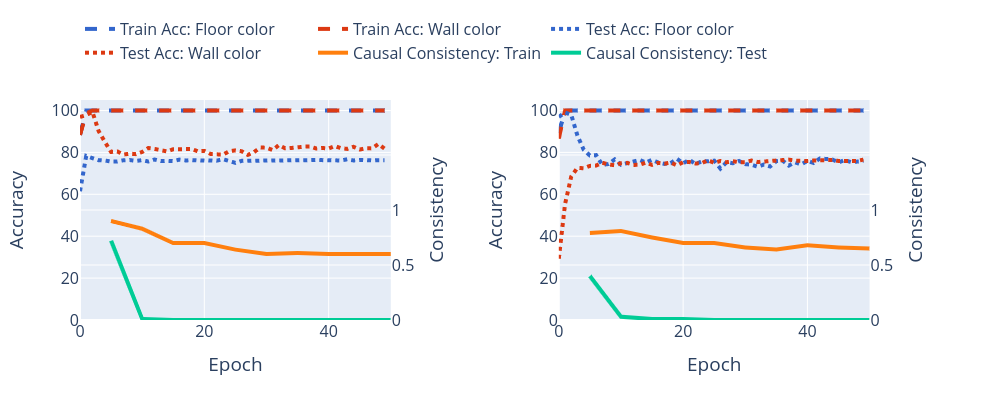}
    \caption{Effect of overfitting on causal consistency of \ourmethod{}.\label{fig:overfit}\vspace{-0.5cm}}
\end{figure}

\vspace{3pt}
\noindent\textbf{Effect of Overfitting:} Here we study the effect of overfitting on the \emph{causal consistency} between the representations. We hypothesize that as the representation learning process overfits, the \emph{causal consistency} on the validation features will drop. Fig.~\ref{fig:overfit} shows the results of this experiment. We observe that after overfitting, the \emph{causal consistency} drops for features from both the training and test set. However, the former still retain some \emph{causal consistency}.


\section{Conclusion}
This paper sought to answer the following questions: \emph{Do learned attribute-specific representations also satisfy the same underlying causal relations? And, if so, to what extent?} To answer these questions, we designed Neural Causal Inference Network (\ourmethod{}) for causal discovery from high-dimensional representations. By bringing together ideas from learning-based supervised, unsupervised causal prediction methods and adversarial debiasing, \ourmethod{} exhibits significantly better causal inference generalization performance. We applied \ourmethod{} to estimate the \emph{causal consistency} between learned representations and the underlying labels in two scenarios, one where the causal relations are known through controlled sampling, and the other where the causal relationships are unknown. Furthermore, we analyzed the effect of overfitting and training epochs on the \emph{causal consistency}. Our experimental results suggest that learned attribute-specific representations \emph{indeed} satisfy the same causal relations between the corresponding attribute labels under controlled scenarios and with high causal strength.

Causal analysis of learned representations is a novel, challenging, and important task. Our work presents a solid yet preliminary effort at answering the questions raised in this paper. Our work is limited from the perspective that there exist many potentially interesting and related aspects of this problem that we did not explore here. From a technical perspective, we foresee two limitations of our work. (1) Unlike the unsupervised methods like RECI, \ourmethod{} needs to be retrained if the dimensionality of the representation changes. (2) \ourmethod{} and the baselines exhibit poor causal inference performance on data with weak causal relations. As the causal relation gets weaker, it is increasingly difficult to distinguish it from the no causal association case. Furthermore, our data generating process does not afford explicit control over the causal strength. 

\noindent\textbf{Acknowledgements:} This work was performed under the following financial assistance award 60NANB18D210 from U.S. Department of Commerce, National Institute of Standards and Technology.

{\small
\bibliographystyle{ieee_fullname}
\bibliography{egbib}
}

\section*{Supplementary Material}

In this supplementary material, we include,
\begin{enumerate}
    \item Precise description and definition of causal consistency in Section~\ref{sec:causal-consistency}
    \item Experimental results of causal consistency on CelebA Face Dataset in Section~\ref{sec:celeba}.
    \item An ablation experiment on effect of the adversarial loss on the performance of \ourmethod{} in Section~\ref{sec:ablation}.
    \item Additional experimental results analyzing the effect of factors such as representation dimensionality and network architecture for learning the representations on \emph{causal consistency} in Section~\ref{sec:discussion}.
    \item Details of the process for generating the synthetic representation for training \ourmethod{} and the baselines in Section~\ref{sec:process}.
    \item Details of the process for generating images with causally associated attributes in Section~\ref{sec:causal-images}.
    \item Details of facial attribute annotation on the CASIA dataset used for our experiments in Section~\ref{sec:casia-annotations}.
\end{enumerate}

\subsection*{1. Definition of causal consistency\label{sec:causal-consistency}}
Datasets are divided into subsets. Causal consistency is the ratio of subsets whose causal relation between representations matches that of the labels, with higher values representing higher consistency. Further, we compute average causal consistency (and confidence intervals) across a small interval $K$ (ten) of epochs after representation learning has converged. Overall, $\mbox{Causal consistency} = \frac{1}{K}\sum_{k=1}^K\frac{\# \mbox{consistent subsets}}{\# \mbox{subsets}}$.

\subsection*{2. Causal consistency of CelebA\label{sec:celeba}}
We also conduct causal inference on representations learned on the CelebA dataset. Specifically, we experiment on the case where causal relations between labels are \emph{unknown}. Similar to the experiments on the CASIA dataset, we chose smiling and narrow eyes as the two attributes to investigate, train and validate the attribute predictors on 10,000/10,000 randomly sampled images using a ResNet-18 architecture. We also apply the entropic causal inference method~\cite{kocaoglu2016entropic} to estimate the causal relation between labels and finding that smiling is a cause of narrow eyes. Table \ref{tab:celeb_all} shows the causal inference results of \ourmethod{} and two baseline. \ourmethod{} exhibits strong \emph{causal consistency} in the correct causal direction. Due to the challenge of selecting a score threshold (see Section 6 of main paper for details) for RECI that generalizes beyond the training data, it classifies all sample as no causal relation. However, if we set the  threshold to 0 and let RECI only infer causal and anti-causal direction, the majority samples will also be inferred as the same directions with labels, which shows that in this case, the causal relation between the features is indeed consistent with that of the labels.
\begin{table}[htbp]
  \centering
  \caption{Causal consistency on CelebA.}
    \begin{tabular}{lccc}
    \toprule
          & NCINet & RECI  & NCC \\
    \midrule
    Causal Consistency & 0.82  & 0.00     & 0.01 \\
    \bottomrule
    \end{tabular}%
  \label{tab:celeb_all}
\end{table}

\subsection*{3. Ablation: Effect of ARL \label{sec:ablation}}

\vspace{- 5pt}

\begin{figure}[ht]
    \centering
    \includegraphics[scale=0.4,trim={0 0 0 1.2cm}]{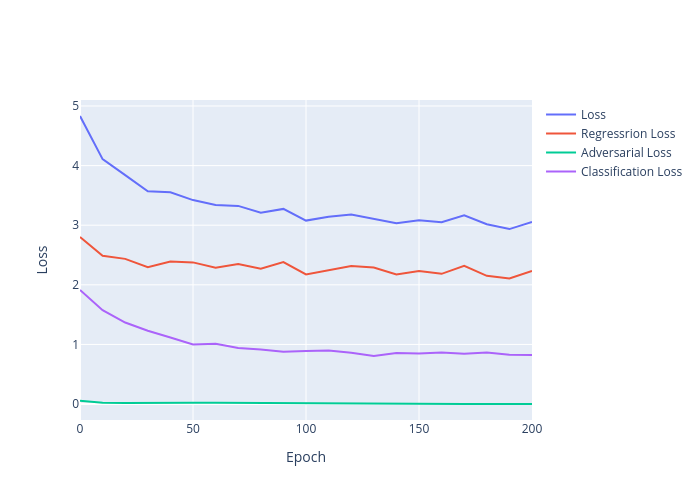}
    \caption{Different components of training loss}
    \label{fig:loss}
\end{figure}

To investigate how adversarial loss contributes to \ourmethod{}, we test three different $\lambda$ values in $Loss = L_C + L_R + \lambda L_A$ and present their generalization results on high-dimensional synthetic data . Table \ref{tab:ablation_weight} shows the generalization results of using different adversarial weight. The results indicate that for data generated from different causal functions, the optimal weight $\lambda$ is different. However, even  a small weight of ARL loss could help the model's generalization ability.

Figure \ref{fig:loss} shows different components of training loss. With a wight $\lambda$ associated with the adversarial loss, all losses are roughly of the
same order of magnitude and well balanced.

\begin{table}[htbp]
    \centering
    \caption{Effect of Adversarial Debiasing on Weight (one run) \label{tab:ablation_weight} \vspace{-0.25cm}}
    \scalebox{0.7}{
    \begin{tabular}{lcccccc}
    \toprule
    \ourmethod{} & \multicolumn{1}{l}{Linear} & \multicolumn{1}{l}{Hadamard} & \multicolumn{1}{l}{Bilinear} & \multicolumn{1}{l}{Cubic spline} &\multicolumn{1}{l}{NN} & \multicolumn{1}{l}{Average} \\
    \midrule
        w/o Adv & 66.50 & 80.33 & 89.67 & 70.5 & 67.17 & 74.83 \\
        optimal Adv & 66.67  & 80.50 & 90.17 & 71.00  & 68.33 & 75.33 \\
        $\lambda$=0.5& 66.67 & 79.67 &89.83 & 70.83 & 68.33& 75.06 \\
        $\lambda$=2&66.67 & 79.83 & 89.67 & 70.83 &68.33 & 75.06\\
        $\lambda$=10&65.00 & 80.50 &90.17 & 71.00 &68.17 & 74.96\\
    \bottomrule
    \end{tabular}}
\end{table}

\begin{figure}[ht]
    \centering
    \includegraphics[scale=0.34]{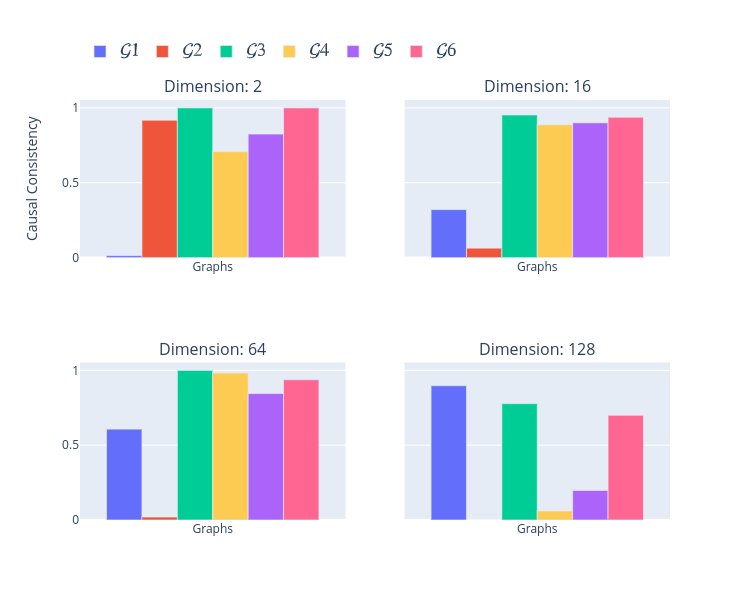}
    \caption{Causal consistency and feature dimension}
    \label{fig:dims}
\end{figure}

\subsection*{4. Discussion\label{sec:discussion}}
\noindent\textbf{Effect of Representation Dimensionality:}

To investigate the effect of representation dimensionality on the inherent causal relations, we evaluate \emph{causal consistency} across different representation dimensionalities on the CASIA WebFace dataset. We set different number of dimensions for the layer before the last linear classifier in the attribute predictor, and extract representations from models that are trained to convergence. Figure \ref{fig:dims} shows the \emph{causal consistency}. We observe that there is slight degradation in the \emph{causal consistency} as the number of dimensions increases, especially at 128 dimensions. However, a more careful and controlled experiment is necessary in order to gain a deeper understanding on the role of representation dimensionality on \emph{causal consistency}.

\begin{figure*}[t]
    \centering
    \includegraphics[scale=0.2]{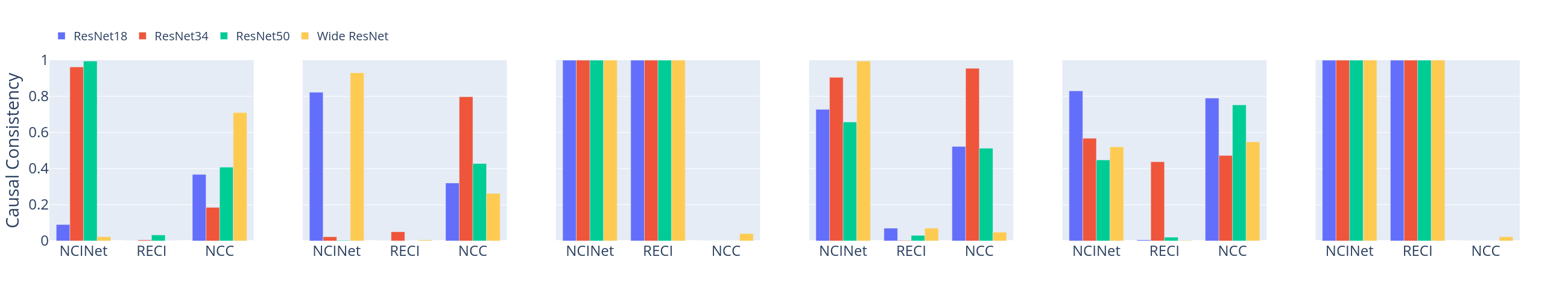}
    \caption{Effect of Architecture and Model Size. From left to right, the plots represent the causal relations encoded by $\mathcal{G}_1$ to $\mathcal{G}_6$.}
    \label{fig:architecture}
\end{figure*}

\vspace{3pt}
\noindent\textbf{Effect of Architecture :} Here we seek to understand if the network architecture has an effect on the causal relations between learned attributes. Therefore, we use four different architecture, including ResNet18, ResNet34, ResNet50 and WideResNet as the attribute predictor for Casia Dataset. Figure~\ref{fig:architecture} shows the causal consistency for multiple network architectures. The results indicates that changes in network architecture have a larger impact on $\mathcal{G}_1$ and $\mathcal{G}_2$, while providing more stable results on other graphs.%

\begin{table}[t]
    \centering
    \caption{Sample complexity ablation. We used $m=100$ for experiments in paper. (one run) \label{tab:sc}}
    \scalebox{0.75}{
    \begin{tabular}{lcccccc}
        \toprule
        & \multicolumn{1}{c}{Linear} & \multicolumn{1}{c}{Hadamard} & \multicolumn{1}{c}{Bilinear} & \multicolumn{1}{c}{Cubic Spline} & \multicolumn{1}{c}{NN} & \multicolumn{1}{c}{Average} \\
        \midrule
        m=10 &    56.50   &   37.00  & 30.67      &   34.00    & 33.67 & 38.36  \\
        m=100 & 66.67  & 80.50 & 90.17 & 71.00  & 68.33 & 75.33 \\
        m=1000 & 58.83 &  81.83   &    90.17   &  70.33     & 66.33 & 73.49 \\
        \bottomrule
    \end{tabular}}
    \end{table}

\vspace{3pt}
\noindent\textbf{Effect of Sample Complexity :} We also study the effect of sample complexity. We set different sample size $m$ and verify the generalization
performance. As shown  in Table \ref{tab:sc}, as sample size increases the results generalize better but plateau with a certain size of sample complexity. The results indicates that to infer the causal relation an adequate number of pairs are needed for each sample.

\begin{table*}[htbp]
  \centering
  \caption{Leave-one-function out accuracy (\%) on different causal functions of different runs.\label{tab:runs}}
    \scalebox{1.0}{
    \begin{tabular}{lcccccccc}
    \toprule
    Methods && \multicolumn{1}{c}{Linear} & \multicolumn{1}{c}{Hadamard} &  \multicolumn{1}{c}{Bilinear} & \multicolumn{1}{c}{Cubic Spline} & \multicolumn{1}{c}{NN} && \multicolumn{1}{c}{Average} \\
    \cline{1-1} \cline{3-7} \cline{9-9} \\
    ANM~\cite{hoyer2009nonlinear}   && 31.87 $\pm$ 1.55 & 32.49 $\pm$ 2.31& 32.94 $\pm$ 0.72 & 33.66 $\pm$ 2.69& 33.08 $\pm$ 1.15 && 32.81 $\pm$ 1.68 \\
    Bfit~\cite{kalainathan2019causal}  && 34.89 $\pm$ 2.01& 54.76 $\pm$ 1.03 & 53.69 $\pm$ 1.70 &\textbf{77.79 $\pm$ 2.40} & 38.26 $\pm$ 1.32 && 51.88 $\pm$ 1.70 \\
    NCC~\cite{lopez2017discovering}   && 52.64 $\pm$ 2.79 & 83.93 $\pm$ 1.55 & 85.66 $\pm$ 1.76 & 77.03 $\pm$ 1.42 & 56.56 $\pm$ 1.37 && 71.16 $\pm$ 1.78 \\
    RECI~\cite{blobaum2018cause}  && 42.73  $\pm$  1.46 & \textbf{89.66  $\pm$  1.50 }& \textbf{92.02  $\pm$  1.01}& 71.49  $\pm$ 0.79 & 60.23 $\pm$  2.15 && 71.43 $\pm$ 1.38\\
    \midrule
    \ourmethod{} && \textbf{64.16  $\pm$  2.33} & 81.13  $\pm$  0.70 & 89.73 $\pm$ 0.71 & 71.33  $\pm$ 0.33 &  \textbf{69.53  $\pm$ 0.94 } && \textbf{75.17 $\pm$ 1.00}\\
    \bottomrule
    \end{tabular}}%
\end{table*}
\vspace{3pt}
\noindent\textbf{Results of Multiple Runs:} To evaluate the stability and effectiveness  of different methods, we run all baselines for five times in the leave-one-function-out generalization experiment, and present their mean accuracy and standard deviation. Specifically, in each run, we generate five different testing datasets for each causal function. The results, shown in Table~\ref{tab:runs}, indicate that \ourmethod{} have a more stable result comparing with other baselines.

\begin{table}[t]
    \centering
    \begin{subtable}[b]{0.5\textwidth}
    \centering
    \caption{Causal consistency on 3Dshape with standard deviation \label{tab:3d_ci}}
    \scalebox{0.6}{
    \begin{tabular}{lcccccc}
    \toprule
    & $\mathcal{G}_1$    & $\mathcal{G}_2$    & $\mathcal{G}_3$   & $\mathcal{G}_4$  & $\mathcal{G}_5$   & $\mathcal{G}_6$ \\
    \midrule
    NCINet &0.89 $\pm$ 0.01 & 0.94 $\pm$ 0.00 & 0.83 $\pm$ 0.02   &0.86 $\pm$ 0.04 & 0.99 $\pm$ 0.01 & 0.79 $\pm$ 0.03 \\
    RECI  & 0.05 $\pm$ 0.00 & 0.21 $\pm$ 0.02  & 0.85 $\pm$ 0.00 & 0.90 $\pm$ 0.00 & 0.90 $\pm$ 0.00 & 1.00 $\pm$ 0.00 \\
    NCC &  0.53 $\pm$ 0.02 &  0.46  $\pm$ 0.02  & 0.00 $\pm$ 0.00 &0.50  $\pm$ 0.00 & 1.00 $\pm$ 0.00 & 0.00 $\pm$ 0.00\\
    \bottomrule
    \end{tabular}}
    \end{subtable}
    
     \begin{subtable}[b]{0.5\textwidth}
    \centering
    \caption{Causal consistency on Casia with standard deviation \label{tab:casia_ci}}
    \scalebox{0.6}{
    \begin{tabular}{lcccccc}
    \toprule
    & $\mathcal{G}_1$    & $\mathcal{G}_2$    & $\mathcal{G}_3$   & $\mathcal{G}_4$  & $\mathcal{G}_5$   & $\mathcal{G}_6$ \\
    \midrule
    NCINet & 0.09 $\pm$ 0.09 & 0.82 $\pm$ 0.11   &1.00 $\pm$ 0.00 & 0.63 $\pm$ 0.09 & 0.82 $\pm$ 0.06 & 1.00 $\pm$ 0.00\\
    RECI  & 0.00 $\pm$ 0.00 & 0.00 $\pm$ 0.01  & 1.00 $\pm$ 0.00 & 0.00 $\pm$ 0.00 & 0.01 $\pm$ 0.02 & 1.00 $\pm$ 0.00 \\
    NCC &  0.36 $\pm$ 0.15 &  0.32  $\pm$ 0.01  & 0.00 $\pm$ 0.00 &0.42  $\pm$ 0.09 & 0.74 $\pm$ 0.05 & 0.00 $\pm$ 0.00\\
    \bottomrule
    \end{tabular}}
    \end{subtable}
    
\end{table}

\vspace{3pt}

\noindent\textbf{Standard Deviation:} 
Table~\ref{tab:3d_ci} and ~\ref{tab:casia_ci} show mean and standard deviation (specific numbers of Figure 4 in main paper) over the small interval of epochs after representation learning has converged on the 3d shape and Casia datasets. As can be observed that causal consistency of NCINet, from one epoch to the other is very stable, which is comparable to unsupervised method.

\subsection*{5. Synthetic Causal Representation Generating Process \label{sec:process}}
\begin{table*}[ht]
  \centering
  \caption{Generative Model for Synthetic Causal Representations \label{tab:causal-functions}}
    \scalebox{0.8}{
    \begin{tabular}{lccccc}
    \toprule
    Causal functions & \multicolumn{1}{c}{Linear} & \multicolumn{1}{c}{Hadamard} &  \multicolumn{1}{c}{Bilinear} & \multicolumn{1}{c}{Cubic spline} & \multicolumn{1}{c}{NN} \\
    \midrule
    \multirow{2}{*}{w/o Confounder} & \multicolumn{5}{c}{$\bm{w} \sim \sum_{k=1}^K \pi_k\mathcal{N}(\bm{\mu}_k,\bm{\Sigma}_k)$ and $\bm{x} = g_x(\bm{w})+\bm{\epsilon}$} \\
    & $\bm{y} = \bm{Ax}+\bm{\epsilon}$ & $\bm{y} = \bm{A}(\bm{x}\odot \bm{x})+\bm{Bx}+\bm{\epsilon}$ & $\bm{y} = \bm{x^TAx}+\bm{\epsilon}$ & $\bm{y} = Spline(\bm{x})+\bm{\epsilon}$ & $\bm{y} = MLP(\bm{x})+\bm{\epsilon}$ \\
    \midrule
    \multirow{2}{*}{w/ Confounder} & \multicolumn{5}{c}{$\bm{w} \sim \sum_{k=1}^K \pi_k\mathcal{N}(\bm{\mu}_k,\bm{\Sigma}_k)$ $\quad$ $\bm{z} = g_z(\bm{w})+\bm{\epsilon}$ $\quad$ $\bm{x} = g_x(\bm{z})+\bm{\epsilon}$} \\
    & $\bm{y} = \bm{A}\bm{\tilde{z}}+\bm{\epsilon}$ & $\bm{y} = \bm{A}(\bm{\tilde{z}}\odot\bm{\tilde{z}}) +\bm{B}\bm{\tilde{z}}+\bm{\epsilon}$ & $\bm{y} = \bm{\tilde{z}}^T\bm{A}\bm{\tilde{z}}+\bm{\epsilon}$ & $\bm{y} = Spline(\bm{x})+Spline(\bm{z})+\bm{\epsilon}$ & $\bm{y} = MLP(\bm{\tilde{z}})+\bm{\epsilon}$ \\
    \bottomrule
    \end{tabular}}%
    \begin{tablenotes}
    \footnotesize
    \item $\bm{\tilde{z}}$ indicates concatenation of  $\bm{x}$ and $\bm{z}$.
    \end{tablenotes}
\end{table*}

The following steps are the detailed data generation process. In this illustration, we taking the case of $X$ being the cause variable for example:
\begin{itemize}

 \item \textbf{Generating initial cause data}: we first sample initial data $W$ from a mixture of Gaussian distributions, and then generate synthetic representation $X$ through a causal function: $X=f(W)+\bm{\epsilon}$.
 
 \item \textbf{Generating ground truth label}: Randomly select one of the first six scenarios in Figure 2 of main paper, and assign the corresponding label to $l$.
 
 \item \textbf{Generating high-dimensional causal relation}: Randomly select one of the five high-dimensional causal function to establish causal relation from cause to effect: $Y=f(X)+\bm{\epsilon}$.
 
 \item \textbf{Confounder Cases}: In the cases which involves confounder $Z$ (e.g., $\mathcal{G}_4$), we first establish the causal relation of $Z \rightarrow X$  , and then establish the causal relation of $X,Z\rightarrow Y$: $Y=f(X,Z)+\bm{\epsilon}$. In the cases where $X$ and $Y$ have no causal relation (i.e. $l=0$), if it involves confounder $Z$, we  establish the causal relation of $Z \rightarrow X$  and $Z \rightarrow Y$,if not, we leave $X$ and $Y$ as their initial values.
\end{itemize}

The five high-dimensional causal functions are specified in Table \ref{tab:causal-functions}, with both w/o confounder and w/ confounder cases. For linear and quadratic functions, we directly multiply the cause variable with coefficient matrices in their form. For Bilinear function, we apply a bilinear transformation to the cause variable. For cubic spline function, we follow \cite{lopez2017discovering}, applying a cubic Hermite spline function. We draw $k$ knots from $\mathcal{N}(0, 1)$, where $k$ is drawn from RandomInteger(5, 20). For Neural Networks function, we apply multilayer perceptrons with hidden layers and numbers of hidden neurons drawn from RandomInteger(0, 3) and RandomInteger(8, 20). For each function, its parameters (e.g., $\bm{A}$, $\bm{B}$ or MLP weights) are drawn at random from $\mathcal{N}(0,1)$ for each data sample. The noise terms $\bm{\epsilon}$ are sampled from Gaussian(0, $v$), where $v \sim$ Uniform(0, 0.1). After each operation, including data initialization and causal relation establishment, the data will be normalized to zero mean and unit variance. Note that for initial data generating, we also apply same causal function as high-dimensional causal relation generating.

\subsection*{6. Generating Images with Causally Associated Attributes\label{sec:causal-images}}

As mentioned in Section 7 of the main paper, the image generating process contains two phases. In the first phases, we  sample labels with six causal relations of Figure 2 in main paper. We first  build  Bayesian Network with hand-designed conditional probability tables of six causal graphs, and then conduct Gibbs Sampling to get attribute labels with known causal relation. The goal of the second phase is to sample images using the labels with known causal relation. For example, in 3D Shapes Dataset, we select attribute floor hue and wall hue as the attribute X and Y in six causal graphs. Then we sample images according to the labels with known causal relations, that is, we select images whose attribute floor hue and wall hue are same with the sampled labels, while we keep other attributes random. For each image, we also randomly add one of three  types of noise, Gaussian, Shot, or Impulse.
\begin{figure*}[h]
\includegraphics[scale=0.8]{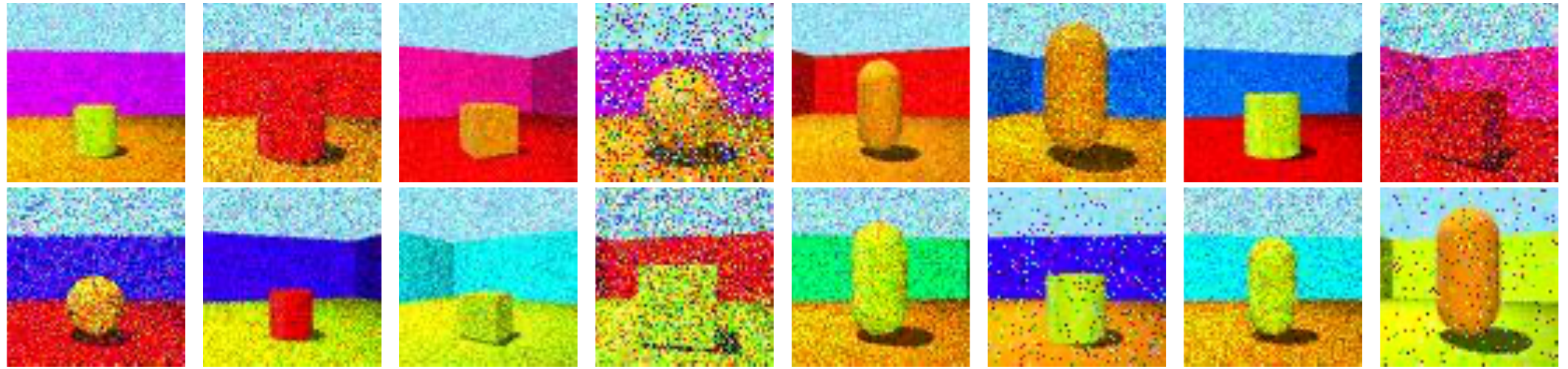}
\caption{Sample images generated from the 3D Shapes dataset with known causal relations. \label{fig:3d-shapes-exmaples}}
\end{figure*}
Figure~\ref{fig:3d-shapes-exmaples} shows examples of images generated from the 3D Shapes dataset. Similarly for facial dataset CelebA and Casia Dataset, we also apply same strategy to sample images using labels with known causal relationship from original dataset.

\subsection*{7. Facial Attribute Annotations\label{sec:casia-annotations}}
Progress in causal discovery methods for computer vision has been hampered by the lack of a large-scale dataset annotated with different underlying causal relations. We posit that existing datasets such as CelebA \cite{liu2015deep}, which has annotations of multi-label attributes in the form of binary labels, is inadequate for causal discovery for a couple of reasons. First, a majority of the images for each attribute are highly imbalanced towards one of the two classes. And more importantly, we observed that a majority of the binary labels are very close to being independent of each other. As such, it may not accurately reflect the causal relations in the real-world and are for the most part are unsuitable as an evaluation benchmark.

To overcome this hurdle we adopt the CASIA-Webface \cite{yi2014learning} dataset, a large public face dataset with 10,575  people and 494,414 images in total, for our experiments. Since this dataset is designed for face verification and recognition problems, only identity annotation is available. Therefore, we augment this dataset with manual annotations of multiple facial attributes (see Table \ref{tab:statistic} for details). The annotated attributes\footnote{The choice of attributes and labels for each may arguably still not fully reflect the real-world. Nonetheless, we believe this dataset could be a valuable resource for causal analysis task.} include: color of hair, visibility of eyes, type of eye wear, facial hair, whether mouth is open, smiling or not, wearing a hat, visibility of forehead, and gender. The annotations for this dataset will be made publicly available to the research community.\footnote{The onus of obtaining the actual images will still remain with the respective research groups.}

The attributes were chosen to be objectively as unambiguous as possible while spanning a range of semantic properties with a variety of causal  relationships amongst them as shown in Figure 1 of main paper. For example, smiling could be a cause of mouth being open because smiling might result in an open mouth. Or, wearing a hat could be a cause for affecting the visibility of forehead, since hats may cause occlusions on people's forehead. Moreover, gender could also causally affect facial hair, because females do not have facial hair in most cases.
\begin{table*}[ht]
  \centering
  \caption{CASIA-WebFace facial attributes, corresponding categories, and sample statistics.\label{tab:statistic}}
  \scalebox{0.5}{
    \begin{tabular}{llllllllllllllllllllllllll}
    \toprule
    \multicolumn{2}{c}{Color of Hair} &       & \multicolumn{2}{c}{Eyes} &       & \multicolumn{2}{c}{Eye Wear} &       & \multicolumn{2}{c}{Facial hair} &       & \multicolumn{2}{c}{Forehead} &       & \multicolumn{2}{c}{Mouth} &       & \multicolumn{2}{c}{Smiling} &       & \multicolumn{2}{c}{Wearing a hat} &       & \multicolumn{2}{c}{Gender} \\
    \midrule
    red   &  12,337      &       & closed & 18,047       &       & none  &   424,128    &       & none  &   364,076    &       & partially visible &   126,219    &       & open  &   215,556    &       & no    &    221,170   &       & no    &  424,659     &       & female &  209,402\\
    gray  & 17,050      &       & open  &  425,185     &       & eyeglasses &   17,805    &       & beard  &   1,763    &       & visible &   297,555    &       & wide open &   16,717    &       & yes   &   231,890    &       & yes   &   28,401    &       & male  & 243,658 \\
    bald  & 13,239      &       & not visible &   9,828    &       & sunglasses &  11,127     &       & mustache &   21,525    &       & fully blocked &  29,286     &       & closed &  220,787     &       &       &       &       &       &       &       &       &  \\
    blonde &   85,848    &       &       &       &       &       &       &       & goatee &   2,613    &       &       &       &       &       &       &       &       &       &       &       &       &       &       &  \\
    black &   158,761    &       &       &       &       &       &       &       & beard and mustache &   48,025    &       &       &       &       &       &       &       &       &       &       &       &       &       &       &  \\
    brown &  144,523     &       &       &       &       &       &       &       & mustache and goatee &   15,058    &       &       &       &       &       &       &       &       &       &       &       &       &       &       &  \\
    not visible &  21,302     &       &       &       &       &       &       &       &       &       &       &       &       &       &       &       &       &       &       &       &       &       &       &       &  \\
    \bottomrule
    \end{tabular}}
\end{table*}

\subsection*{8. Gradient of Closed-Form Solution}
In order to find the gradient of the kernel ridge regressor of adversary, we rewrite the loss function of adversary as:

\begin{equation}
\begin{split}
\label{e1}
    L_A &= -\|\bm{y}_f - \hat{\bm{y}}_f\|^2_2 = -\|\bm{y}_f - \bm{K}\left(\bm{K} + \beta \bm{I}\right)^{-1}\bm{y}_f\|_2^2\\&= -\|(\bm{I}-\bm{K}\left(\bm{K} + \beta \bm{I}\right)^{-1})\bm{y}_f\|\\&=-\|P_{\bm{K}}\bm{y}_f\|
\end{split}
\end{equation}

Then from \cite{golub1973differentiation}, letting $\theta$ be arbitrary scalar
element of encoder, we have 

\begin{equation}
\label{e2}
    \frac{1}{2}\frac{\partial \|P_{\bm{K}}\bm{y}_f\|^2}{\partial \theta}=\bm{y}_f^TP_{\bm{K}^\perp}\frac{\partial \bm{K}}{\partial \theta}\bm{K}^{\dagger}\bm{y}_f,
\end{equation}
where $\bm{K}^\perp$ is the orthogonal complement of $\bm{K}$, and

\begin{equation}
    \big[\frac{\partial \bm{K} }{\partial \theta}\big]_{ij} = \begin{cases} \nabla^T_{\bm{z}_i}\big([\bm{K} ]_{ij}\big)
\nabla_\theta({\bm{z}_i})
+\nabla^T_{\bm{z}_j}\big([\bm{K} ]_{ij}\big)
\nabla_\theta({\bm{z}_j}), & i\le n\\0, & \text{else}.
    \end{cases}
\end{equation}
Equation \ref{e2} can be directly used to obtain the gradient of objective function in \ref{e1}.
The gradient of ridge regressor from unsupervised branch can be derived in same way by simply replacing the kernel matrix $\bm{K}$ with linear one.

\end{document}